\documentclass[runningheads]{llncs}
\usepackage[T1]{fontenc}
\usepackage{graphicx}
\usepackage{times}
\usepackage{latexsym}

\usepackage{framed}
\usepackage{booktabs}
\usepackage{multirow}
\usepackage{tabularx}
\usepackage{makecell}
\usepackage{url}
\usepackage{inconsolata}
\usepackage{amsmath}
\usepackage{caption}

\usepackage{graphicx} 
\usepackage{float} 
\usepackage{amssymb}
\usepackage{subcaption}			

\usepackage[utf8]{inputenc}

\usepackage{microtype}

\usepackage{inconsolata}
\begin{document}
\title{High-Quality Data Augmentation for Low-Resource NMT: Combining a Translation Memory, a GAN Generator, and Filtering}
%
\titlerunning{High-Quality Data Augmentation for Low-Resource NMT}
%
\author{Hengjie Liu\inst{1} \and
Ruibo Hou \inst{2} \and
Yves Lepage\inst{3}}
\authorrunning{H. LIU et al.}
%
\institute{Waseda University, IPS ~
 \email{yo4c5ama@toki.waseda.jp}\\
\and
Waseda University, IPS ~
 \email{houruiboc@akane.waseda.jp}\\
\and
Waseda University, IPS ~
 \email{yves.lepage@waseda.jp}}
\maketitle              
\vspace*{-1\baselineskip}
\begin{abstract}

Back translation, as a technique for extending a dataset, is widely used by researchers in low-resource language translation tasks.
It typically translates from the target to the source language to ensure high-quality translation results. 
This paper proposes a novel way of utilizing a monolingual corpus on the source side to assist Neural Machine Translation (NMT) in low-resource settings. 
We realize this concept by employing a Generative Adversarial Network (GAN), which augments the training data for the discriminator while mitigating the interference of low-quality synthetic monolingual translations with the generator. 
Additionally, this paper integrates Translation Memory (TM) with NMT, increasing the amount of data available to the generator. 
Moreover, we propose a novel procedure to filter the synthetic sentence pairs during the augmentation process, ensuring the high quality of the data.

\end{abstract}

\section{Introduction}

There are many languages around the world that are on the brink of extinction.
Machine translation plays a crucial role in preserving endangered languages. 
However, a common challenge in this endeavor is the scarcity of parallel corpora available online, which hinders the development of effective translation systems.

In recent years, the rise of Neural Machine Translation (NMT) \cite{NIPS2014_a14ac55a} make to a technological leap.
At the beginning of the research of NMT, researchers used deep neural networks, especially Recurrent Neural Network (RNN) and Long Short-term Memory Networks (LSTM), to better capture language context and complex structures. 
Moreover, some researchers transfered Convolution Neural Network (CNN) in the classfication task of Natural Language Processing (NLP) \cite{sharma2020sentimental}.
However, today, most of the researchers design their models based on transformer \cite{NIPS2017_3f5ee243}.
The transformer model, based on attention mechanisms, allows more flexible focus on different parts of the input sentence, further improving translation accuracy.
It also set new benchmarks for most other tasks in the field of NLP. 

Low-resource translation \cite{koehn-knowles-2017-six} refers to the situation where machine translation encounters a shortage of parallel corpora for training models. 
This commonly occurs for languages with limited linguistic resources or smaller speaker populations, leading to difficulties in achieving effective translation performance.

A Translation Memory (TM) is a repository that archives pairs of source sentences along with their corresponding translations. 
Recent studies have validated the beneficial impact of TM on enhancing NMT models. 
This enhancement has been demonstrated through various approaches, including concatenating both the source and target side of the TM \cite{wang-lepage-2022-translation} and the target side of the TM with the source input \cite{xu-etal-2020-boosting}, encoding TM and source input separately \cite{he-etal-2021-fast}, and leveraging TM in Non-autoregressive machine translation models \cite{xu-etal-2023-integrating}. 
However, research \cite{cai-etal-2021-neural} indicates that integrating TM into the NMT model does not yield improvements over the baseline NMT model when applied to the same task in a low-resource setting.


In low-resource translation tasks, researchers often seek to expand parallel corpora to enhance model performance. 
Back-translation is a popular data augmentation technique among researchers.
However, it has some drawbacks. 
The sentences generated through back-translation frequently differ from natural language sentences, potentially disrupting and degrading the translator's training process. 
A proposed method utilizes a monolingual corpus on the target side for back-translation \cite{sennrich-etal-2016-improving}. 
This approach involves back-translating target language sentences to the source language and adding these synthetic sentences to the parallel data, ensuring high-quality target-side translations. 
Building on this work, \cite{fadaee-monz-2018-back} investigated various aspects of back-translation and proposed several sampling strategy variations targeting difficult-to-predict words using prediction losses and word frequencies. 
Additionally, iterative back-translation to expand low-resource corpora was explored by \cite{hoang-etal-2018-iterative}, who proposed a method to generate increasingly better synthetic parallel data from monolingual data to train neural machine translation systems.

Currently, when using back-translation to handle low-resource translation tasks, researchers typically use monolingual corpora from the target language to ensure that the synthetic sentences does not interfere with the model in training process. 
However, when the target language is a low-resource language, the available monolingual corpora are extremely limited.
Therefore, this paper aims to propose a method that utilizes monolingual corpora from the source language to improve low-resource translation tasks. 
In devising this method, the paper leverages the structural characteristics of Generative Adversarial Networks (GAN) to achieve this goal.


GAN architectures have not been extensively applied in NMT tasks, primarily due to the inherent instability of the GAN training process, which is especially evident in NMT. 
A simple GAN architecture for NMT was proposed by \cite{pmlr-v95-wu18a}, suggesting the use of a Convolution Neural Network (CNN) as a discriminator and an NMT model as a generator. 
Given the exceptional performance of the Transformer model in various NLP tasks, \cite{yang-etal-2018-improving} proposed employing a Transformer as a generator. 
To address the training instability issues of traditional GANs in NMT, \cite{zhang-etal-2018-bidirectional} introduced a novel Bidirectional Generative Adversarial Network (BGAN) for NMT. 
However, even when other researchers employ GANs to tackle NMT tasks, they primarily focus on model design and selection rather than effectively harnessing the adversarial nature of the GAN framework.

In this paper, integrating the findings of previous work, we propose harnessing data augmentation of low-resource via GAN integrating TM. Here are our contributions: 
\begin{itemize}
    \item We enable the utilization of a source-side monolingual corpus without interfering with the translator's performance by employing synthetic data.
    \item We integrate a TM into the generator of GAN, enlarging the amount of data which can be trained by generator. 
    \item We design a novel filtering procedure to ensure the high-quality of the translations.   
\end{itemize}

\section{Methodology}

\subsection{Integrating TM into NMT}
\vspace*{-2\baselineskip}
\begin{figure*}[ht]
	\centering
	\resizebox{0.5\columnwidth}{!}{
	\includegraphics[scale=1]{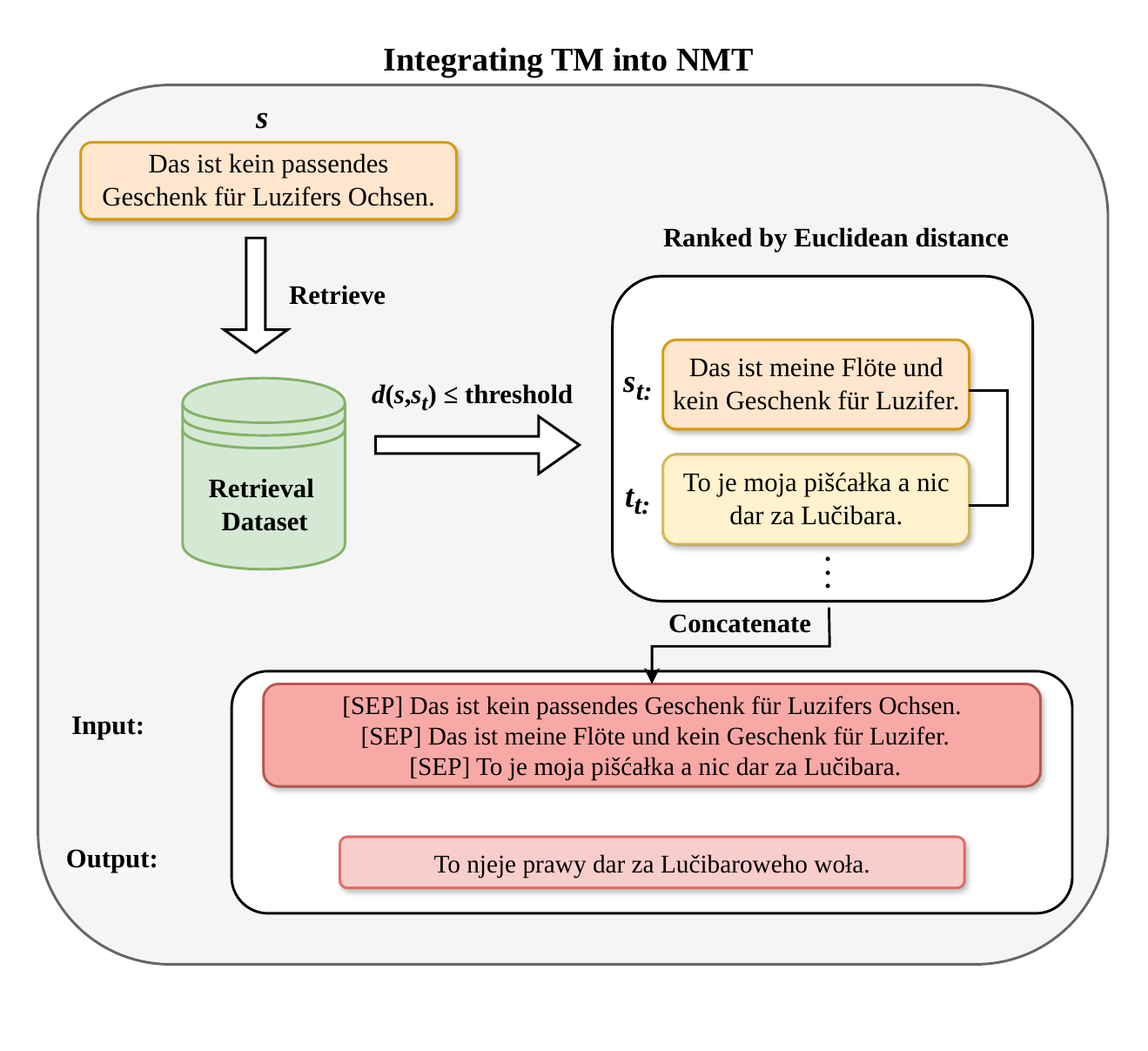}
	}
	\vspace*{-1\baselineskip}
	\caption{Process of integrating TM into NMT, where $d(s,s_t)$ means the Euclidean distance between sentence vectors of source input $s$ and a source sentence $s_t$ in TM. The input consists of $s$, $s_t$ and $t_t$. The corresponding output is the translation $t$ of $s$.}     
	\label{fig:Integrating TM}
\end{figure*}
\vspace*{-1\baselineskip}
We employ the retrieval method to find sentences $s_t$ in the TM that are semantically similar to the source input $s$, along with their corresponding target sentences $t_t$. 
Here, we introduce a threshold for the Euclidean distance between sentence vectors to ensure the quality of the retrieved TM. 
To ensure the quality of the retrieved synthetic sentences that will be introduced in Section \ref{subsection:Back-Translation Leveraging Monolingual Corpus on Source Side}, we will discuss how to control generation in Section \ref{subsection:Back-Translation Leveraging Monolingual Corpus on Source Side} and how to filter high qulaity sentences in Section \ref{subsection:High-quality Filtering}.
Additionally, considering the limited number of sentences in the low-resource language TM, we avoid setting this threshold too high or too low. 
In subsequent experiments, the threshold is set to 0.5 \cite{xu-etal-2020-boosting}.
\begin{equation}
	\sqrt{\sum_{i=1}^{n}[s_i - (s_t)_i ]^2 }  \le 0.5
\end{equation}

Sentence pairs $(s_t,t_t)$ with a distance below this threshold will be concatenated with $s$ using the following format: '[SEP] $s$ [SEP] $s_t$ [SEP] $t_t$', where [SEP] denote separators. 
The process of integrating TM into the input is presented in Figure \ref{fig:Integrating TM}.

\vspace{\baselineskip}
\noindent
\textbf{Retrieval Method} ~ Suppose that our retrieval dataset is a set of source and target sentence pairs $(s_t,t_t)$, which is the same set that would be used as a training dataset for an NMT system. 
For a given source input, denoted as $s$, it is treated as the query to do semantical retrieval in retrieval dataset. 
The corresponding semantically similar source sentence, denoted as $s_t$ in the TM, serves as the associated value. 
Consequently, obtaining the corresponding target sentence $t_t$, and the sentence pair $(s_t,t_t)$ becomes a straightforward process. 

Firstly, we input each sentence in the TM into a pre-trained sentence embedding model \cite{reimers-2019-sentence-bert} to generate sentence vectors. 
Subsequently, we calculate the similarity score between two sentences by computing the Euclidean distance between their respective sentence vectors.

To facilitate rapid retrieval between the input vector representation and the corresponding vector of sentences in the TM, we employ FAISS toolkit \cite{johnson2019billion}.

\subsection{Back-Translation Leveraging Monolingual Corpus on Source Side}
\label{subsection:Back-Translation Leveraging Monolingual Corpus on Source Side}
\vspace*{-2\baselineskip}
\begin{figure*}[ht]
	\centering
	\resizebox{0.7\columnwidth}{!}{
	\includegraphics[scale=1]{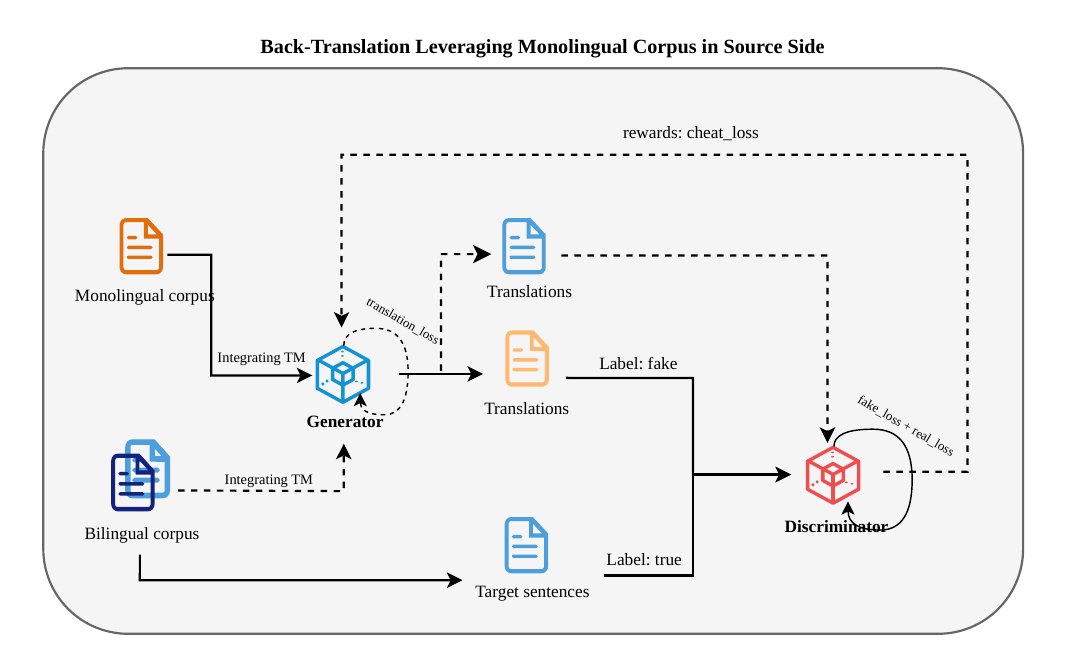}
	}
	\vspace*{-1\baselineskip}
	\caption{The way utilizing monolingual corpus in source side enhance NMT task. The Generator is a vanilla Transformer model. The Discriminator is a fusion neural network model design by ourselves.}     
	\label{fig:GAN_processing}
\end{figure*}
\vspace*{-1\baselineskip}
In this paper, integrating the findings of previous work and the current research landscape of NMT, we use a vanilla Transformer as a generator, $G$, for complex translation tasks, to generate synthetic data in the target language, i.e., the low-resource language in our settings. 
To build this generator $G$, we design a fusion neural network as a discriminator $D$, composed of parallel BiLSTM and CNN, handling a binary classification task.
As \cite{NIPS2014_5ca3e9b1,pmlr-v95-wu18a,zhang-etal-2018-bidirectional} explained, $D$ and $G$ play the following two-player minimax game with value function $V(G, D)$:
\vspace*{-0.5\baselineskip}
\begin{equation}
	\begin{split}
		\underset{G}{\min}~\underset{D}{\max}~V(D, G)= \mathbb{E}_{(x, y) \sim P_{data}(x, y)}[\log D(x, y)] \\
	   +~\mathbb{E}_{(x_{1}, y_{1}^{\prime}) \sim P_{G}(x, y)}[\log (1-D(x, y^{\prime}))]
	\end{split}
	\label{GAN}
\end{equation}

In the Equation \eqref{GAN}, $(x, y)$ is a sentence pair from the bilingual corpus, $(x_{1},y_{1}^{\prime})$ is a source sentence in the monolingual corpus combined with the target sentence generated by $G$. 
$P_{data}$ represents the real data distribution and $P_{G}$ denotes the generator distribution. 
In this way, $D$ continually learns how to distinguish real and natural sentences and not so natural sentences generated by $G$. 
Meanwhile, $G$ strives to produce the most natural sentences to deceive $D$.

\vspace{\baselineskip}
\noindent
\textbf{Training Strategy} ~ Figure \ref{fig:GAN_processing} illustrates our process of performing back-translation by leveraging a monolingual corpus on the source side. 
We employ a vanilla Transformer model as the GAN generator. 
The monolingual corpus, which is similar to but distinct from the bilingual corpus, is utilized to supply negative samples to the discriminator. 
In the training process, we first reform the monolingual corpus by integrating TM as the input of the generator. 
After obtaining the translations from the generator, we label these synthetic translation sentences from the monolingual corpus as fake sentences. 
We separate the natural target sentences from the bilingual corpus as ground-truth sentences. 
Our discriminator is trained by combining these two batches of sentences, labeled as unnatural and natural (fake and true in GAN parlance), separately. 
The output of the discriminator from fake sentences is used to calculate the loss value with $0$, while the output from true sentences is used to calculate the loss value with $1$. 
We add these loss values together with the same weight and propagate the gradient back to the discriminator to guide the classification task.

On the generator side, two loss functions are employed to improve performance. 
Since this is essentially still a translation task, we use the original loss function of the translation task. 
This differs somewhat from a classical GAN model. 
The loss value is derived from training the generator using the bilingual corpus. 
Additionally, we feed the translations generated from the bilingual corpus into the discriminator. 
The discriminator's outputs are then used to compute the cross-entropy loss with $1$, which is fed back to the generator. 
The generator combines this loss with the translation task loss using a smaller weight. 
This approach aims to assist in improving the translation model.
In this way, the generator continuously produces more natural sentences to deceive the discriminator. 

By using monolingual corpora in this manner, we observe that synthetic sentences derived from monolingual translation not only avoid interfering with the translation model but also enhance the discriminator's capability along with real sentences. 
In turn, the discriminator helps improve the performance of the generator. 
Thus, we achieve the goal of using monolingual corpora on the source side to improve the model's translation ability without negatively impacting the translation model.

\begin{figure*}[ht]
	\centering
	
	\includegraphics[scale=0.5]{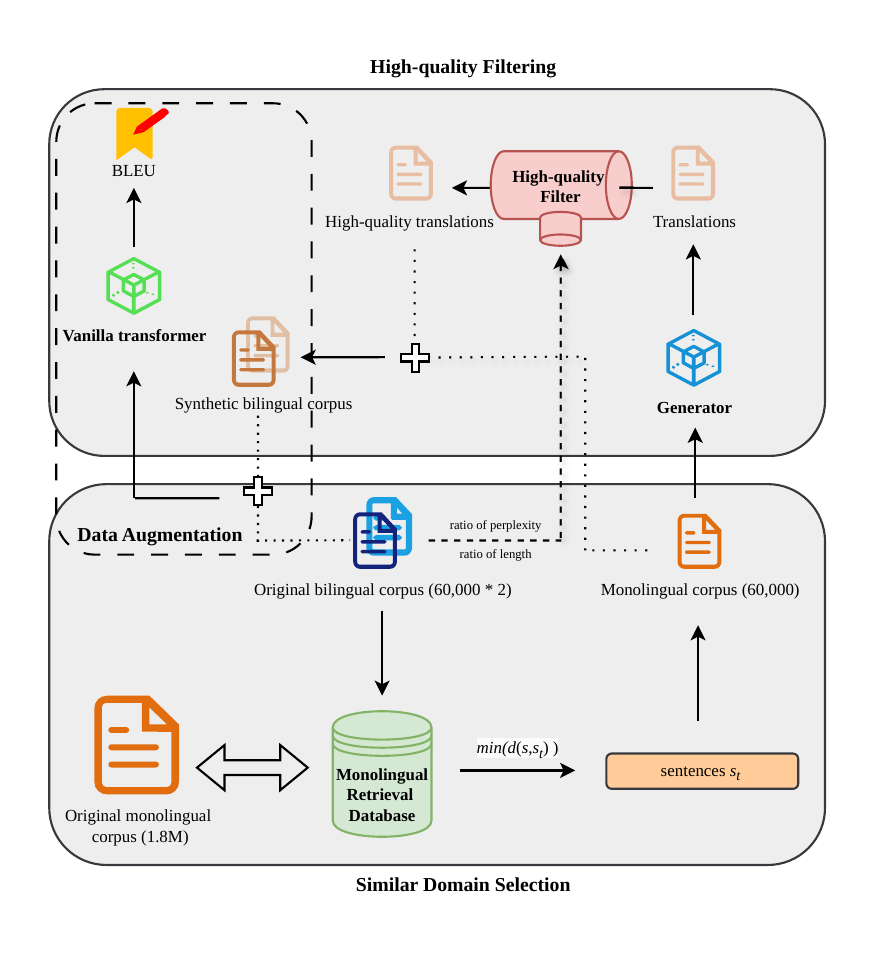}
	\vspace*{-1\baselineskip}
	\caption{A process for filtering high-quality translations. We filter both the source sentences and the target sentences. We calculate features from natural language corpora (the ratio of sentence length and perplexity) to serve as filtering criteria. This allow us to filter translations that are more fit with natural language sentences. Finally, we validate the effectiveness of the translation results using standard data augmentation experiments.}     
	\vspace*{-2\baselineskip}
	\label{fig:Domain_filtering_augmentation}
\end{figure*}

\subsection{High-quality Filtering}
\label{subsection:High-quality Filtering}
To demonstrate the effectiveness of the low-resource sentences translated by our generator, we utilize data augmentation experiments for verification. 
However, in a standard data augmentation process, the synthetic sentences may vary in quality, and we cannot ensure that all sentences in the synthetic corpus are of high quality. 
Based on this, this paper proposes an effective high-quality filtering process,  as illustrated in Figure \ref{fig:Domain_filtering_augmentation}.

\vspace{\baselineskip}
\noindent
\textbf{Similar Domain Selection} ~ Since the source language is high-resource, we have a large amount of data to choose from. 
To initially ensure the quality of the translated sentences, we perform similar domain selection on the original monolingual corpus. 
First, we build a retrieval database based on the original monolingual corpus. 
For each source language sentence $s$ in the original bilingual corpus, we select the closest sentence $s_{t}$ from the retrieval database. 
These $s_{t}$ sentences form the monolingual corpus to be translated after the similar domain selection.

\vspace{\baselineskip}
\noindent
\textbf{High-quality Filtering Method} ~ To filter high-quality sentences in the translation, we designed a high-quality filtering. 
As described by Gale-Church alignment algorithm \cite{gale-church-1993-program}, in a natural bilingual parallel corpus, the length ratio between source and target sentences typically varies around a fixed value, and these ratios generally follow a Gaussian distribution. 
This paper posits that, besides length, other features of parallel sentence pairs also adhere to such a specific relationship. 
We use a language model toolkit \cite{heafield-2011-kenlm} to train $N$-gram models for both the source (de) and target (hsb) languages and compute their perplexities. 
Preliminary experiments reveal that the distribution of perplexity ratios for parallel sentence pairs also approximates a Gaussian distribution. 
Hence, we use the interval [mean - standard deviation; mean + standard deviation] of the ratio of these two features (length and perplexity) in a natural parallel corpus as the filtering criterion for high quality.
For each translation sentences, we calculate the length and perplexity ratios between the source sentence and itself. 
Only if both ratios fall within the filtering criterion, the translation is considered high-quality and passes through the filter.

\vspace{\baselineskip}
\noindent
\textbf{Data Augmentation} ~  We conduct a standard data augmentation experiment to demonstrate the effectiveness of our filtering method. 
We combined the filtered high-quality translations with their source sentences to form a synthetic bilingual corpus. 
Then, we merged this synthetic bilingual corpus with the original bilingual corpus and used them to train a vanilla Transformer model. 
Finally, we calculated relevant metrics to prove the effectiveness of the translations from our generator.

\section{Setup \& Evaluation}
\subsection{Dataset}
We use the German-Upper Sorbian (de-hsb) parallel corpus from WMT20\footnote{\url{https://www.statmt.org/wmt20/unsup_and_very_low_res/}} as the original bilingual corpus.
The training sets of the dataset consist of 60,000 parallel sentences, while both the validation and test sets contain 2,000 sentences each.
We choose German monolingual corpus from WMT14\footnote{\url{https://www.statmt.org/wmt14/training-monolingual-news-crawl/}} as the original monolingual corpus.
This German monolingual corpus consist of 1,879,765 German sentences. 
We apply filtering and do not utilize all of the data.
Since we will try to understand what is the necessary and sufficient amount of additional data to attain the best performance.
The statistics of the corpora used are shown in Table \ref{table:Statistics of the Corpora}.
\vspace*{-1\baselineskip}
\begin{table}[ht]
	\begin{center}
    \resizebox{0.7\columnwidth}{!}{
		\begin{tabular}{llccr}
			\toprule
			    \multirow{2}*{Corpus} & \multirow{2}*{Language} & \multirow{2}*{\# sentences} & \multicolumn{2}{c}{Sentence length}  \\
				\cline{4-5}
				~                     &            ~          &                ~             & in words             & in characters \\
			\midrule
				\multirow{2}*{bilingual} &  German (de)       &  \multirow{2}*{60,000 $\times$ 2}   & 12.1 ± 6.9           & 83.4 ± 51.7 \\
						~			  &  Upper Sorbian (hsb)  &                ~             & 10.7 ± 6.3           & 71.6 ± 45.4\\
			\midrule
				monolingual  		&       German (de)		  &            1,879,765         &	15.4 ± 9.2			&	108.5 ± 64.8 \\
			\bottomrule
		\end{tabular}
    }	
		\caption{Statistics of the corpora used}
		\label{table:Statistics of the Corpora}
	\end{center}
\end{table}
\vspace*{-5\baselineskip}
\subsection{Models \& Evaluation}
Given the outstanding performance of Transformers in NLP tasks, especially NMT, we choose a vanilla Transformer model as our generator.
Its encoder and decoder each have 8 heads and 12 layers in total. We set the embedding dimension to 512 and the dropout rate to 0.15. 
We employe a warm-up strategy, after which the learning rate decays with the inverse square root schedule. 
On the discriminator side, we use a parallel combination of CNN and BiLSTM. 
The CNN had a convolutional kernel size of $16 \times 1$, while the BiLSTM had a hidden layer size of 256. 
Both the discriminator and generator use an Adam optimizer.

All of our models and training processes were implemented using PyTorch \cite{paszke2017automatic}. 
The FAISS toolkit \cite{johnson2019billion} was utilized for similar sentence retrieval, while the kenLM toolkit \cite{heafield-2011-kenlm} was employed for training $N$-gram models. 
All model training was conducted on a single A4500 GPU. 
To evaluate the translation results, we used BLEU \cite{papineni-etal-2002-bleu}, chrF2 \cite{popovic-2015-chrf}, and TER \cite{snover-etal-2006-study} as the evaluation metrics. 
The calculation of these metrics were implemented through the sacreBLEU toolkit \cite{post-2018-call}.

\section{Experimental Results}
\subsection{Performance of Generators}
\label{subsection:Performance of Generators}
The first group of rows in Table \ref{table:Data Augmentation} shows the performance of the generator with different components. 
In this series of experiments, we used a vanilla Transformer model as the baseline. 
This model, serving as the generator, achieved a BLEU score of 37.2 for translation. 
Building upon this baseline, we incorporated different components.

We first integrated TM into the input of the model. 
The results surpassed the baseline by 1.4 BLEU points.
Considering the confidence intervals.
This improvement is not significant, as the intervals for the baseline model (37.2 ± 1.2) and the model with TM integrated (38.6 ± 1.2) overlap (38.2 - 1.2 = 37.0 $<$ 38.4 = 37.2 + 1.2). 

Subsequently, we trained the generator using a monolingual corpus in conjunction with a GAN.
However, we found that the performance improvement was not significant (37.3 - 1.1 = 36.2 $<$ 38.4 = 37.2 + 1.2). 
We believe that this is due to the instability of GAN training, which also confirms that GANs are not well-suited for NMT tasks \cite{zhang-etal-2018-bidirectional}. 
Of course, this is also related to the weight proportions of different loss values during the training process. 
Nevertheless, utilizing GANs was not our primary purpose. 
Our most important goal in employing GANs was to introduce source-side monolingual corpora into the model training process. 

From the metrics, we can observe that when both TM and GAN are integrated into the system, the translation performance improves by 3.7 BLEU points compared to the baseline. 
This improvement is significant, as the intervals for the baseline model (37.2 ± 1.2) and the model with TM and GAN integrated (40.9 ± 1.2) do not overlap (40.9 - 1.2 = 39.7 $>$ 38.4 = 37.2 + 1.2).
These four sets of experiments demonstrate that TM is an effective method for enhancing the translation performance of the model. 
We believe that this is because TM increases the amount of available data for model training and serves as a prompt for the model. 
The GAN structure, on the other hand, enables the utilization of source-side monolingual corpora. 
The synthetic sentences not only do not interfer with the generator but also indirectly facilitate the learning process of the generator through the discriminator.

For comparison with the method proposed by \cite{sennrich-etal-2016-improving}, a reverse translation model (from target to source) was trained in this study (see the 1st row in Table \ref{table:Data Augmentation}).

\begin{table*}[ht]
	\begin{center}
		\resizebox{\linewidth}{!}{
		\begin{tabular}{lccccrrr}
			\toprule
				New sentence 	 &	 Amount of & \multicolumn{3}{c}{Filter with}	 &	\multirow{2}*{BLEU}	&	\multirow{2}*{chrF2}	&	\multirow{2}*{TER}	\\
				\cmidrule(r){3-5} 
				pairs from   & new sentence pairs  &   random   &   perplexity   &   length    &        ~      &       ~       &      ~         \\
			  \midrule
        Vanilla Transformer (reverse side) &       0     &      ~     &       ~        &      ~    &	 39.6 ± 1.1 &	 66.6 ± 0.8	    &	45.1 ± 1.0   \\
        \midrule
				Vanilla Transformer  & \multirow{4}*{0}  &      ~     &       ~        &      ~        &	 \underline{37.2 ± 1.2} &	\underline{64.2 ± 0.8}	&	\underline{46.1 ± 1.1}   \\
				+TM	       &      ~     &       ~        &      ~     &        ~	      &	 38.6 ± 1.2	&	64.5 ± 0.8	&	44.3 ± 1.1   \\
				+GAN	     &      ~     &       ~        &      ~     &        ~        &	 37.3 ± 1.1	&	64.6 ± 0.8	&   45.9 ± 1.0   \\
				+TM +GAN 	 &      ~     &       ~        &      ~     &        ~	      &	 \textbf{40.9 ± 1.2}	&	\textbf{67.6 ± 0.8}	&	\textbf{39.8 ± 1.0}   \\
			  \midrule
				+TM +GAN 	 & 60,000 &      ~     &       ~        &      ~	     &	37.4 ± 1.1	&	64.3 ± 0.8	&	45.6 ± 1.0 	    \\
        		Sennrich's work \cite{sennrich-etal-2016-improving}& 60,000       &      ~      &       ~      &      ~	    &	37.7 ± 1.2	&	 64.8 ± 0.8	 &	 44.8 ± 1.1	    \\
				+TM +GAN 	 & 10,000 & \checkmark &       ~        &      ~	     &	38.2 ± 1.2	&	65.0 ± 0.8	&	45.5 ± 1.1  	\\
				+TM +GAN	 & 40,052 &      ~     &       ~        & \checkmark	 &	38.1 ± 1.1	&	64.7 ± 0.8	&	45.0 ± 1.0 	    \\
				+TM +GAN	 & 19,408 &      ~     &    \checkmark  &      ~       &	38.8 ± 1.2 	&	65.1 ± 0.9	&	44.5 ± 1.1 	    \\
				+TM +GAN 	 & 13,464 & \checkmark &       ~        &      ~   	   &	38.8 ± 1.1	&	65.3 ± 0.8	&	44.4 ± 1.0  	\\
				Sennrich's work \cite{sennrich-etal-2016-improving} & 13,464  &      \checkmark     &       ~        &      ~	     &	39.1 ± 1.2	&	 65.5 ± 0.8	 &	 44.3 ± 1.0	    \\
        +TM +GAN	 & 13,464 &      ~     &    \checkmark  & \checkmark  &	\textbf{40.1 ± 1.3}	&	\textbf{66.2 ± 0.8} 	&	\textbf{43.6 ± 1.1}   	\\
			\bottomrule
		\end{tabular}
		}
		\caption{The table compares the performance of different sentence filtering methods on a NMT and augmentation task. The first group of rows in the table presents the NMT task used to train our generator model and its performance. The second group of rows shows that we used the generator to augment the original bilingual corpus. These augmented bilingual corpora were then used to train a vanilla Transformer model, which was subsequently evaluated on the test set. }
		\label{table:Data Augmentation}
	\end{center}
\end{table*}

\subsection{High-quality Filtering \& Results}
\label{subsection:High-quality Filtering & Results}

As shown in the second group of rows of the Table \ref{table:Data Augmentation}, we used the best performing generator from the first group of rows (combining TM and GAN) for translation and conducted data augmentation experiments following the process illustrated in Figure \ref{fig:Domain_filtering_augmentation}.

Initially, we augmented the original bilingual corpus with the entire monolingual corpus (60,000 German sentences) and their translations without any filtering (see the 6th row in Table \ref{table:Data Augmentation}). 
We found that the results hardly improved. 
After analysis, we concluded that this was due to the gap between the synthetic translation sentences and real natural language. 
Augmenting with an equal number of synthetic sentences as real sentences would interfere with the translation model, which is consistent with our previous analysis. 

Therefore, we reduced the quantity and randomly selected 10,000 synthetic parallel sentence pairs for augmentation (see the 8th row in Table \ref{table:Data Augmentation}). 
As expected, the results showed preliminary improvement, surpassing the baseline by 1.0 BLEU point.

We then changed our filtering strategy to ensure the high quality for the selected sentence pairs.
As introduced in Section \ref{subsection:High-quality Filtering}, we employed three filtering methods using the sentence length ratio, sentence perplexity ratio, and a combination of length ratio and perplexity ratio from the original bilingual corpus as the filtering criterion. 
The experimental results demonstrate that using both length and perplexity ratios as the filtering criterion yields the best performance, surpassing the baseline by 2.9 BLEU points (see row number 2 and 13 in Table \ref{table:Data Augmentation}: $40.1 - 1.3 = 38.8 > 38.4 = 37.2 + 1.2$).

Additionally, to prove that the experimental results are not entirely influenced by the number of sentences, we randomly selected the same number of sentences (13,464) as the best-performing method (see the 11th row in Table \ref{table:Data Augmentation}). 

For Sennrich's method, 13,464 sentences were also randomly selected. In comparison, our method demonstrated better performance (see row number 12 and 13 in Table \ref{table:Data Augmentation}). 
However, the improvement was not significant. 
In the back-translation method employed by Sennrich, the target sentences are natural. 
Therefore, it can be concluded that after our high-quality filtering process, the synthesized target sentences also more closely resemble natural sentences.
The results were lower than those obtained using the two filtering methods. 
Therefore, we conclude that our filtering process is effective.

\begin{figure*}[ht]
	\centering
	\begin{minipage}[c]{0.2\textwidth}
		\centering
		\includegraphics[width=\textwidth]{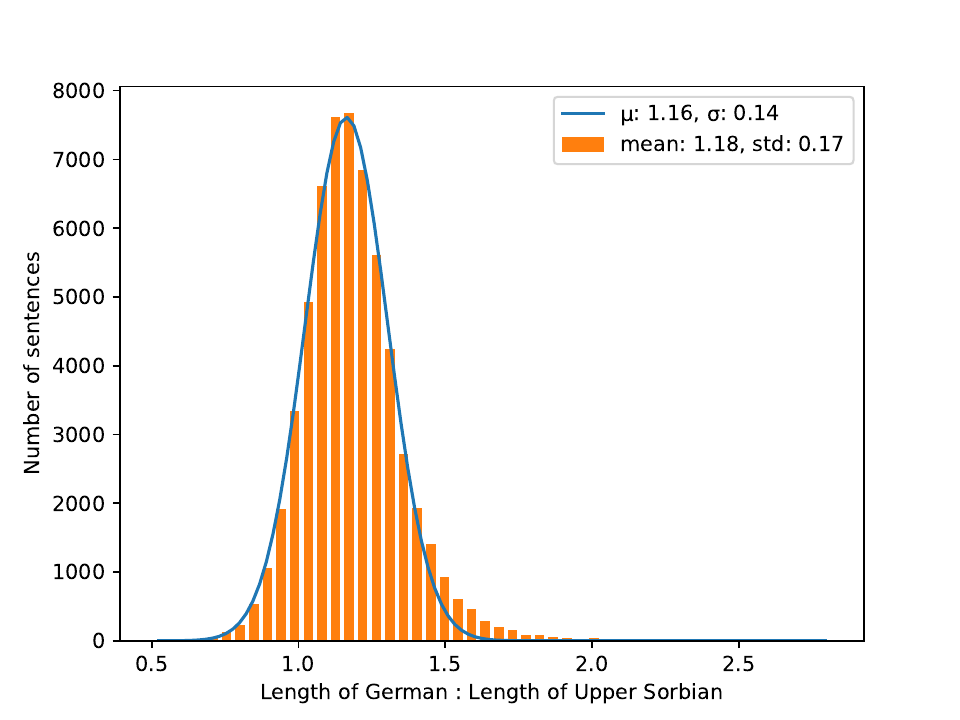}
		\vspace{-1.5\baselineskip}
		\subcaption{}
		\label{fig_filter_1}
	\end{minipage} 
	\begin{minipage}[c]{0.2\textwidth}
		\centering
		\includegraphics[width=\textwidth]{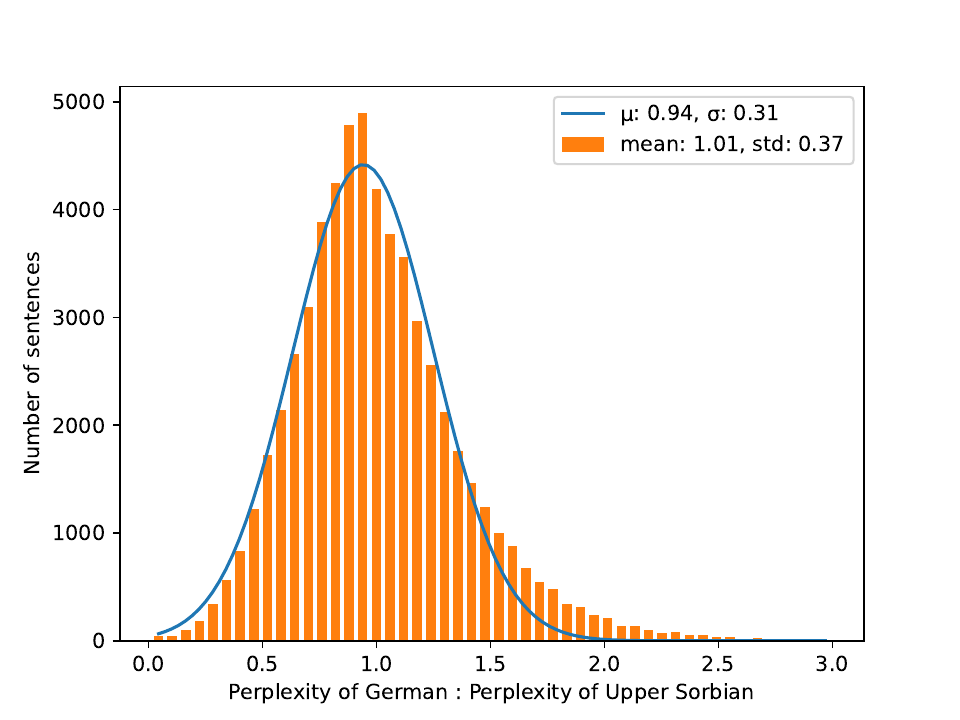}
		\vspace{-1.5\baselineskip}
		\subcaption{}
		\label{fig_filter_2}
	\end{minipage} 
	\begin{minipage}[c]{0.2\textwidth}
		\centering
		\includegraphics[width=\textwidth]{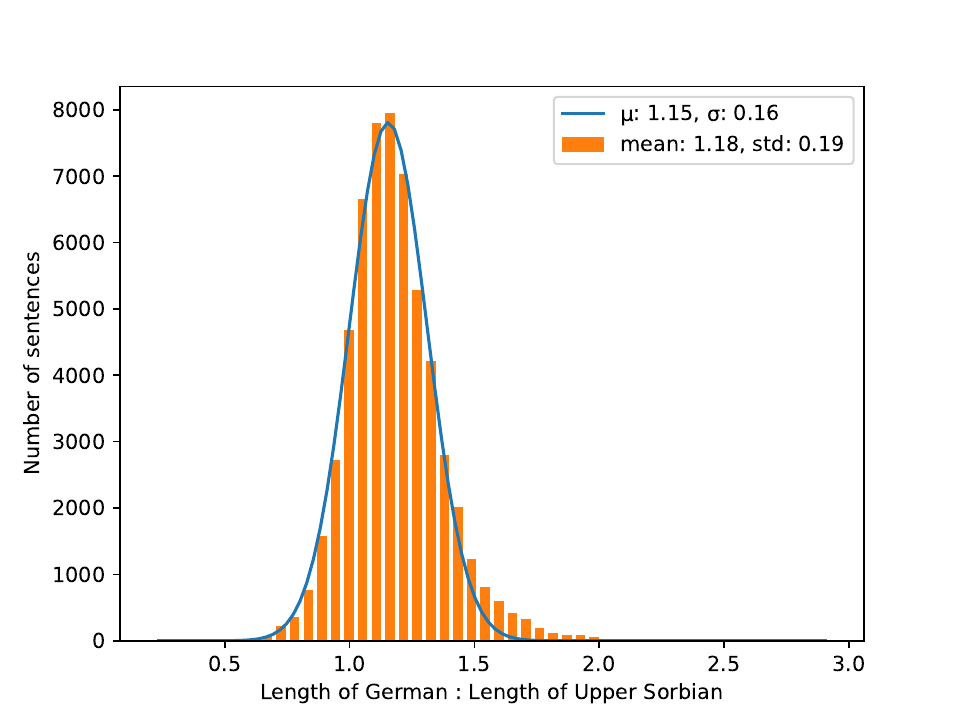}
		\vspace{-1.5\baselineskip}
		\subcaption{}
		\label{fig_filter_3}
	\end{minipage}
	\begin{minipage}[c]{0.2\textwidth}
		\centering
		\includegraphics[width=\textwidth]{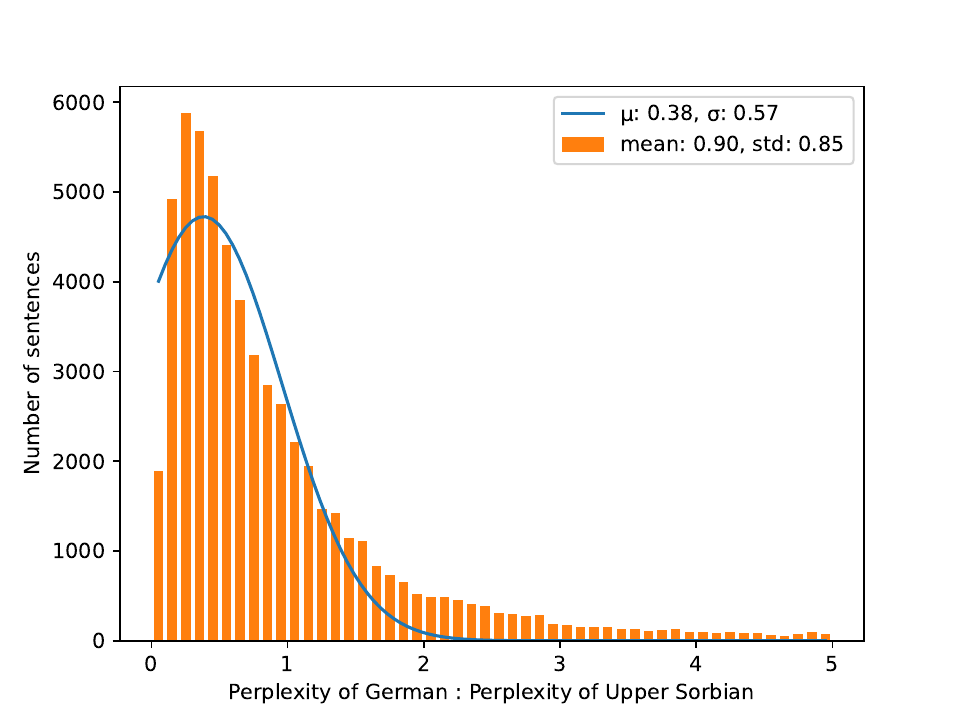}
		\vspace{-1.5\baselineskip}
		\subcaption{}
		\label{fig_filter_4}
	\end{minipage}\\
	\begin{minipage}[c]{0.2\textwidth}
		\centering
		\includegraphics[width=\textwidth]{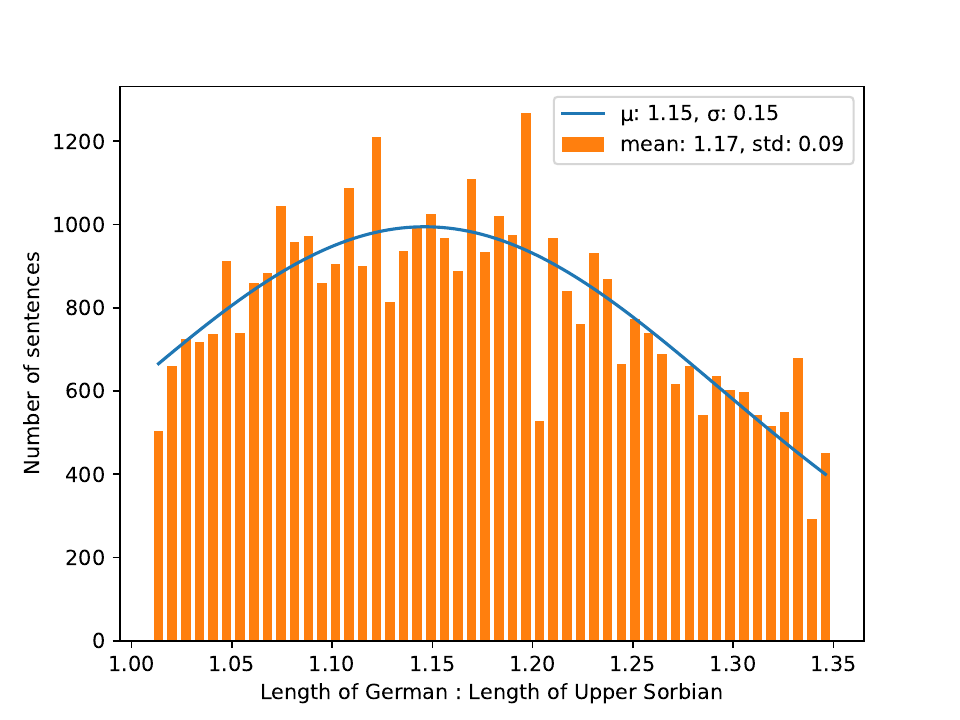}
		\vspace{-1.5\baselineskip}
		\subcaption{}
		\label{fig_filter_5}
	\end{minipage} 
	\begin{minipage}[c]{0.2\textwidth}
		\centering
		\includegraphics[width=\textwidth]{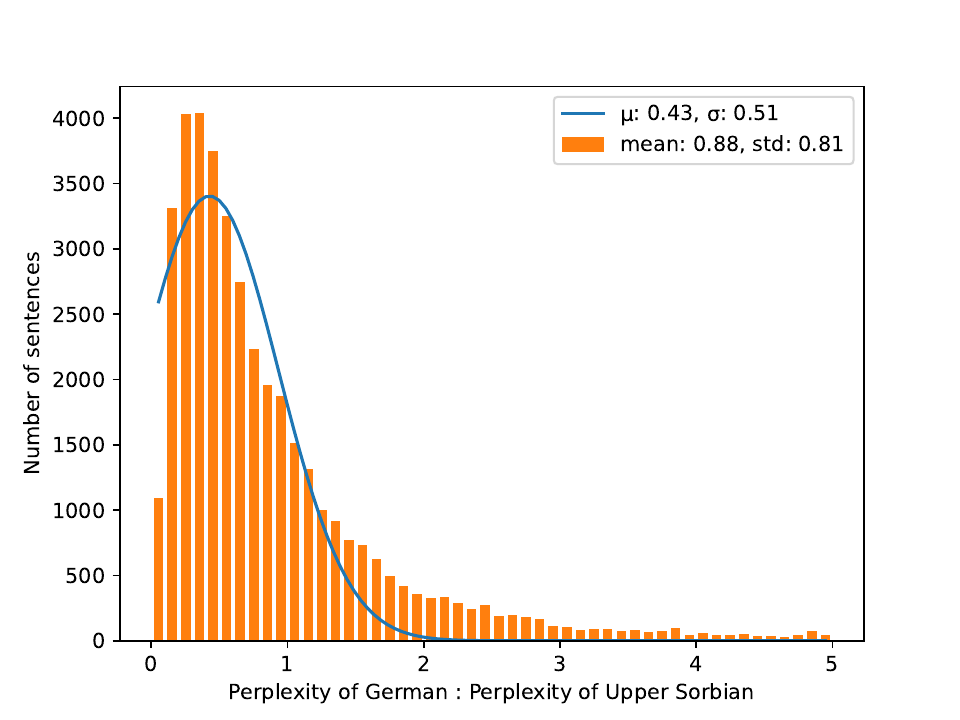}
		\vspace{-1.5\baselineskip}
		\subcaption{}
		\label{fig_filter_6}
	\end{minipage} 
	\begin{minipage}[c]{0.2\textwidth}
		\centering
		\includegraphics[width=\textwidth]{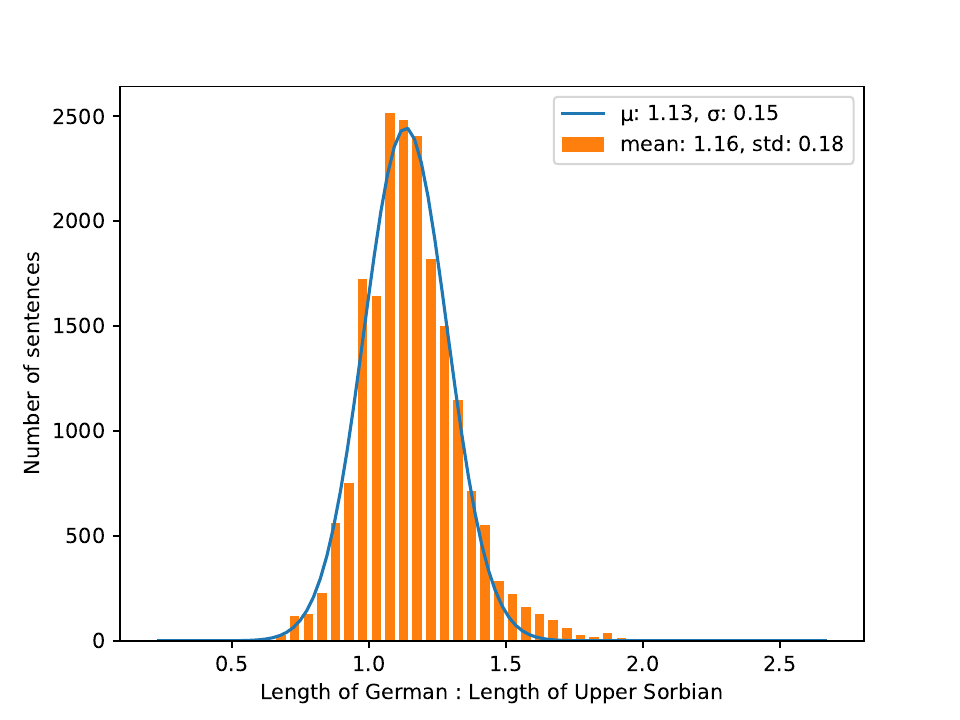}
		\vspace{-1.5\baselineskip}
		\subcaption{}
		\label{fig_filter_7}
	\end{minipage}
	\begin{minipage}[c]{0.2\textwidth}
		\centering
		\includegraphics[width=\textwidth]{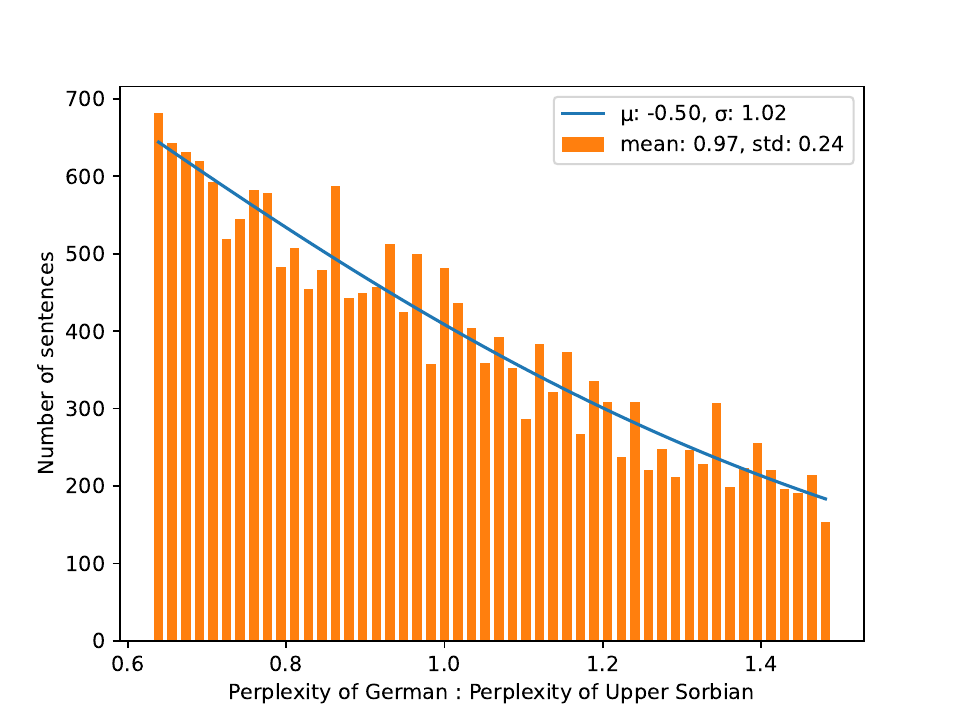}
		\vspace{-1.5\baselineskip}
		\subcaption{}
		\label{fig_filter_8}
	\end{minipage}\\
	\begin{minipage}[c]{0.2\textwidth}
		\centering
		\includegraphics[width=\textwidth]{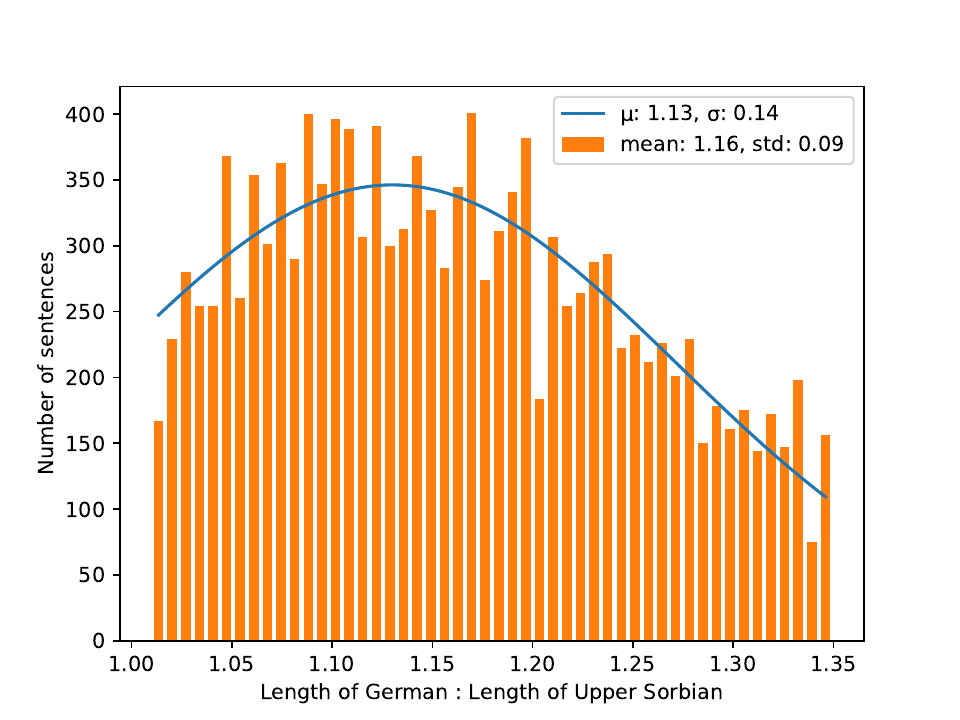}
		\vspace{-1.5\baselineskip}
		\subcaption{}
		\label{fig_filter_9}
	\end{minipage}
	\begin{minipage}[c]{0.2\textwidth}
		\centering
		\includegraphics[width=\textwidth]{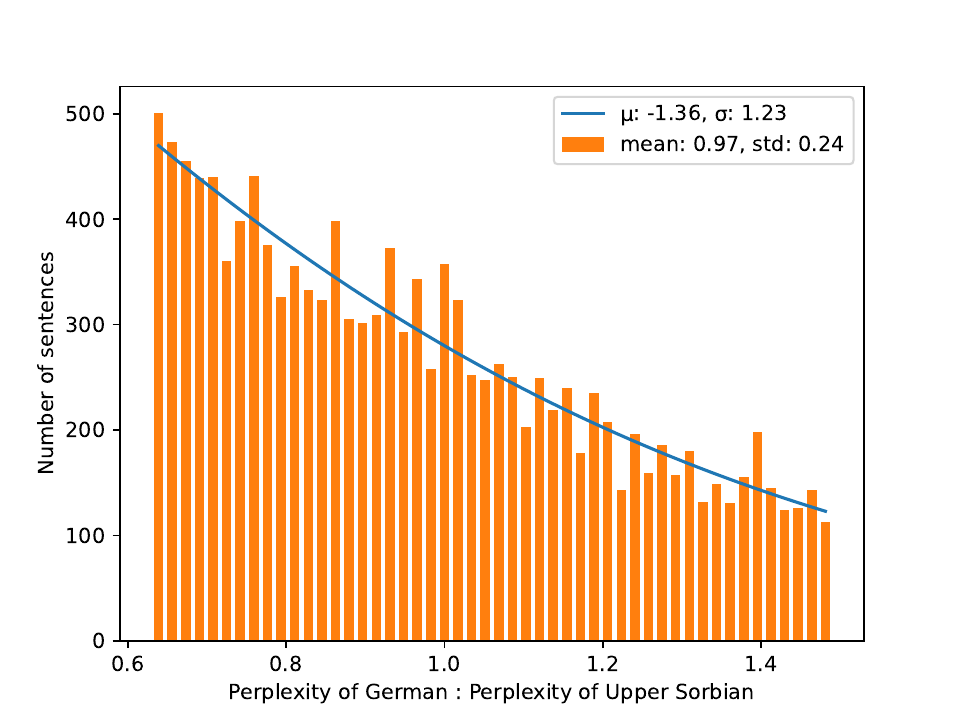}
		\vspace{-1.5\baselineskip}
		\subcaption{}
		\label{fig_filter_10}
	\end{minipage}\\
	\vspace{-1\baselineskip}
	\caption{Figures (a) and (b) illustrate the distribution of the original bilingual corpus in terms of sentence length ratio and perplexity ratio, respectively. In contrast, (c) and (d) depict the distribution of our newly created bilingual corpus. Subsequently, we first used the length ratio of the original bilingual corpus as the filtering criterion (1.18 ± 0.17) to filter the created bilingual corpus, with the corresponding distribution shown in (e) and (f). Figures (g) and (h) present the distribution when using the perplexity ratio as the filtering criterion (1.01 ± 0.37). Finally, (i) and (j) demonstrate the distribution obtained by employing both length and perplexity as the filtering methods.}
	\label{fig_filter}
	\vspace{-1.5\baselineskip}
\end{figure*}

\vspace{\baselineskip}
\noindent
\textbf{High-quality Filtering} ~ Figure \ref{fig_filter} specifically illustrates the distribution of sentence length and perplexity ratios after filtering.
Comparing subfigures \ref{fig_filter_1} and \ref{fig_filter_3}, we observe that the features of the synthetic sentences translated by our generater largely conform to the features of natural sentences in terms of length.
However, there is still a discrepancy in perplexity, as shown in subfigures \ref{fig_filter_2} and \ref{fig_filter_4}. 
Subfigure \ref{fig_filter_4} is shifted to the left overall compared to \ref{fig_filter_2}, which precisely reflects that some of the synthetic sentences (hsb) have higher perplexity, resulting in smaller ratios. 
This hints at some deficiency in the synthesized sentences.
For that reason, subsequently, we used the mean ± standard deviation of the sentence length ratio and perplexity ratio of the sentence pairs in the original bilingual corpus as the filtering criterion (1.18 ± 0.17) and (1.01 ± 0.37), respectively. 
The specific distribution is shown in the Figure \ref{fig_filter}.

\section{Conclusion}
\label{section:Conclusion}
This paper introduced a novel approach to leveraging a monolingual corpus on the source side when the source language was highly resourced, but the target language was low-resourced, to support NMT in low-resource scenarios. 
We realized this concept by employing a GAN, which augmented the training data for the discriminator while mitigating the interference of low-quality synthetic monolingual translations with the generator. 
Furthermore, this paper integrated TM with NMT, increasing the amount of data available to the generator. 
Additionally, we proposed a novel criterion to filter the synthetic sentence pairs in the augmentation process, ensuring the high quality of the data.

However, our approach also has limitations. 
The training process of GANs is unstable and requires a substantial amount of time. 
Moreover, the semantics and other information inherent in natural language are complex and implicit, making GANs not particularly suitable for natural language translation tasks. 
As for TM, it consumes a significant amount of computational resources. 
Regarding our high-quality filtering process, it cannot guarantee that all the filtered sentence pairs are of high quality. 
Future research can delve into these aspects for further improvements.
%
%
%
\bibliographystyle{splncs04}
\bibliography{custom, anthology}
%




\end{document}